\documentclass[letterpaper, 10 pt, conference]{ieeeconf}  

\IEEEoverridecommandlockouts                              

\usepackage{subcaption}

\usepackage{ifpdf}
\usepackage{graphicx}
\ifpdf
\usepackage{amsmath}
\usepackage{autobreak}
\usepackage{hhline}
\usepackage{multirow}
\usepackage{hyperref}
\usepackage{graphicx}
\graphicspath{{Figs/}{Figs/PDF/}}
\else
\graphicspath{{Figs/}}
\fi
\usepackage{hhline}
\usepackage{color}
\usepackage{calc}
\let\labelindent\relax
\usepackage{enumitem}
\usepackage{bm,amsmath}

\usepackage{nomencl}
\usepackage{multirow}
\usepackage{url}
\usepackage{nomencl}
\usepackage{multirow}
\usepackage{gensymb}
\hypersetup{
	colorlinks = true,
	linkcolor = blue,
	citecolor = blue
}
\usepackage{subcaption}

\usepackage[flushleft]{threeparttable}
\usepackage{cite}
\usepackage{times}
\usepackage{epsfig}
\usepackage{graphicx}
\usepackage{amsmath}
\usepackage{amssymb}

\usepackage{xcolor}

\usepackage{textcomp}

\usepackage{wrapfig}
\usepackage{siunitx}
\usepackage{ifthen}

\newcommand{\danielreplace}[2]{%
\ifthenelse{ \equal{#1}{} }{}{\textcolor{olive}{\sout{#1}}}%
\ifthenelse{ \equal{#2}{} }{}{ \textcolor{olive}{#2}}%
}

\usepackage[compact]{titlesec}
\titlespacing{\section}{-1pt}{2ex}{1ex}
\titlespacing{\subsection}{0pt}{1ex}{0ex}
\titlespacing{\subsubsection}{0pt}{0.5ex}{0ex}

\title{\LARGE \bf
TouchRoller: A Rolling Optical Tactile Sensor for Rapid Assessment of Large Surfaces
}

\author{Guanqun Cao$^{*}$, Jiaqi Jiang$^{*}$, Chen Lu, Daniel Fernandes Gomes and Shan Luo
\thanks{$*$ indicates equal contributions.}
\thanks{All the authors are with the smARTLab, Department of Computer Science, University of Liverpool, L69 3BX. Emails:~\{g.cao, jiaqi.jiang, c.lu15, danfergo, shan.luo\}@liverpool.ac.uk.}

}

\begin{document}

\maketitle
\thispagestyle{empty}
\pagestyle{empty}

\begin{abstract}


Tactile sensing is important for robots to perceive the world as it captures the texture and hardness of the object in contact and is robust to illumination and colour variances.
However, due to the limited sensing area and the resistance of the fixed surface, current tactile sensors have to tap the tactile sensor on target object many times when assessing a large surface, i.e., pressing, lifting up and shifting to another region. This process is ineffective and time consuming. It is also undesirable to drag such sensors as this often damages the sensitive membrane of the sensor or the object.
To address these problems, we propose a cylindrical optical tactile sensor named \textit{TouchRoller} that can roll around its center axis.
It maintains being in contact with the assessed surface throughout the entire motion, which allows for measuring the object continuously and effectively.
Extensive experiments show that the TouchRoller sensor can cover a textured surface of $8 cm\times11 cm$ in a short time of \SI{10}{\s}, much more effectively than a flat optical tactile sensor (in \SI{196}{\s}). The reconstructed map of the texture from the collected tactile images has a high Structural Similarity Index (SSIM) of $0.31$ on average, when compared with the visual texture. In addition, the contacts on the sensor can be localised with a low localisation error, \SI{2.63}{\mm} in the center regions and \SI{7.66}{\mm} on average.
The proposed sensor will enable the fast assessment of large surfaces with high-resolution tactile sensing, and also the effective collection of tactile images. 


\end{abstract}


\section{Introduction}
Tactile sensing is one of the key sensing modalities for robots to perceive the world as it conveys important surface information of the contacting objects such as textures and hardness. Such tactile information not only enables robots to have a better understanding of the target objects, but also help robots facilitate dexterous manipulation. Different from cameras providing a global Field of View (FoV) of the scene, a tactile sensor attains local tactile information by interacting directly with the objects, which provides a smaller sensing area. To this end, it requires large-scale samples when evaluating a large surface using a tactile sensor. The data collection is much more costly compared with cameras, and it poses challenges to the design of a tactile sensor that can effectively collect tactile data.


\begin{figure}[h]
\begin{center}
\includegraphics[width=1\linewidth]{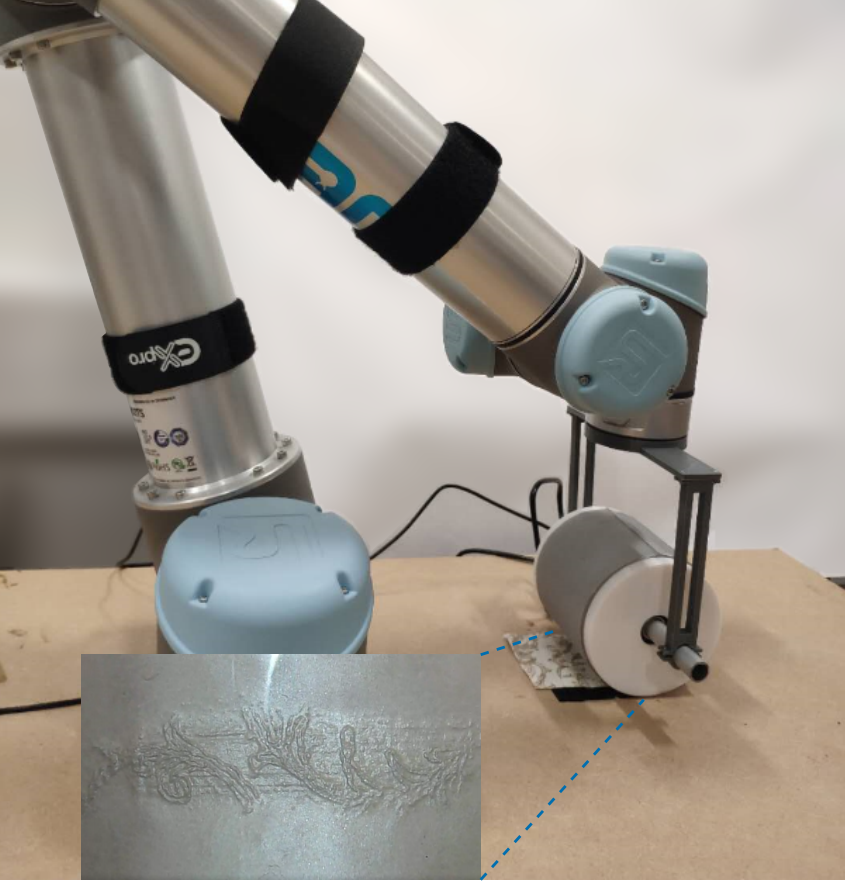}
\end{center}
\caption{Our TouchRoller sensor is rolling over a piece of fabric to collect surface textures, and the collected tactile image of the fabric texture is shown on the bottom left. }
\label{fig:scene}
\end{figure}

To facilitate efficient data collection, a tactile sensor is expected to reduce the ineffective movements that the current tactile sensors perform, e.g., pressing on the objects, lifting up and shifting to another region. Instead, the sensor can adapt an effective motion such as rolling or sliding on the object surface. Moreover, a large sensing area and high sensitivity are also important for the tactile sensor in the general purpose of robotic perception and manipulation. 


Optical tactile sensors use a camera to capture the deformation of a sensitive elastomer on the top of the sensor~\cite{yuan2017gelsight,ward2018tactip}, which offers a high spatial resolution. However, due to the use of soft elastomer, these sensors suffer lack of mobility. The motions of slipping or dragging the sensor on the object's surface result in an unstable contact and a blurred tactile image. Furthermore, the elastomer of the sensor will be easily damaged by the rubbing. As a result, the data collection using such sensors is ineffective and time consuming due to the repeated movements like pressing.

In this paper, we propose a novel cylindrical optical tactile sensor named as \textit{TouchRoller}, as shown in Fig.~\ref{fig:scene}, that can roll over the surface of an object to collect tactile images continuously. When the sensor gets in contact with the object, the elastomer coated on a transparent cylinder distorts to take the geometry of the region in contact, and the embedded camera inside the cylinder will capture this deformation with a high spatial resolution. A sequence of tactile images from different contact locations will be recorded continuously as the sensor rolls over the target object, which can be used to reconstruct the whole object surface effectively. It maintains a stable contact with the assessed surface during rolling, thereby generating clear tactile images and enabling rapid assessment of large surfaces. 

In our experiments, the TouchRoller sensor demonstrates strong capacity of mapping the textures of the object surface efficiently. It can cover a textured surface of $8cm\times11cm$ in 10 s, much quicker than a flat optical tactile sensor (in 196 s). In addition, the sensor can localise contacts around the cylinder with a low localisation error, \SI{2.63}{\mm} in the center regions and \SI{7.66}{\mm} on average.

\section{Related Works}\label{relatedwork}

In the past decades, different designs of tactile sensors based on different sensing principles have been proposed~\cite{luo2017robotic,dahiya2009tactile}, e.g., resistive, capacitive, piezoelectric, magnetism and optical tactile sensors. Among them, optical-based tactile sensors offer superior sensing resolution~\cite{shimonomura2019tactile}, compared to other tactile sensors. Tactile array sensors other than optical ones normally contain tens to hundreds of tactile sensing elements. It would be difficult to further increase the amount due to the constraints of wiring methods. In contrast, optical tactile sensors capture contact information with their built-in cameras, with each pixel seen as a sensing element.



It has been one of the popular methods to use elastic membranes as the tactile skin in the designs of optical tactile sensors. The TacTip sensor~\cite{ward2018tactip} is an open source membrane-based tactile sensor inspired by the sensing mechanism of intermediate ridges in human epidermal layers. The original design~\cite{chorley2009development} has most of its rigid parts 3D-printed despite of the manually molded silicone rubber skin. Arrays of white pins are painted on the skin in a way analogous to the intermediate ridges. Any contact that causes the movement of the pins can be detected by the camera. An upgraded version~\cite{cramphorn2017addition} introduced dual-material 3D-printing to reduce the manufacturing difficulty and product variance. The TacTip sensor has simple manufacturing process and reasonable building cost. In addition, it can be modified for various applications like lesions diagnosis~\cite{winstone2015biomimetic}. However, its measurable area and resolution is limited by the marker-based sensing approach. Furthermore, due to the fixed surface of the sensor, the motions like slipping and dragging will result in an unstable contact and may break the membrane.


The GelSight sensor~\cite{yuan2017gelsight} is another type of membrane-based optical tactile sensor. The sensor consists of a transparent elastomer, an optical camera and a set of LED illumination units. Unlike the TacTip sensor, the transparent elastomer of GelSight sensor is coated by a reflective surface. As a result, the external illumination will not affect the inside space of the sensor. In a typical sensor-object contact scenario, the surface of the membrane will deform according to the target object and capture its geometry. The first GelSight sensor~\cite{johnson2009retrographic} was created to measure the geometry and texture of the target surface. Following the principle of the prototype, multiple upgraded versions of GelSight sensor~\cite{yuan2017gelsight,dong2017improved} have been proposed. The earlier designs of the GelSight sensors are bulky and the sensing area is limited to a small area in the centre of the elastomer.
To improve the sensing area, a finger-shaped sensor named GelTip sensor~\cite{gomes2020geltip} was proposed recently, to enable 360 degree contact detection with full resolution.

However, all of these sensors suffer from the limitation of sensing area and the need of repeated actions like pressing to assess large surface areas. As a result, in the tasks where the target area is much larger than the tactile skin, it would be challenging to acquire tactile data and the data collection is quite time consuming. It is also undesirable for such sensors to perform movements such as slipping and dragging which may lead to the unstable contact and blurred tactile images. In this work, we address these problems by proposing a tactile sensor that has a cylindrical shape to promote the efficiency of data collection. It is able to roll on the object, maintaining the stable contact with the assessed surface to collect tactile features continuously. In ~\cite{shimonomura2019tactile}, the author refers a similar structure of tactile sensor which is illustrated in a workshop. However, we cannot find any publications and scientific analysis about this sensor. In this work, we will discuss about the the fabrication, working mechanism, contact localization as well as surface mapping of the proposed tactile sensor.


\section{The Principle of the TouchRoller Sensor}
\subsection{Overview}
As is illustrated in Fig.~\ref{fig:scene} and Fig.~\ref{fig:shell}, our proposed TouchRoller sensor consists of six main parts: 1) a piece of transparent elastomer, 2) a reflective membrane painted on the elastomer, 3) a transparent cylindrical tube, 4) a camera to capture the tactile features, 5) embedded LEDs to illustrate the space inside the sensor, as well as 6) supporting parts to connect all elements of the sensor.

When the sensor rolls over an object surface, e.g., a piece of fabric, the soft elastomer will deform according to the fabric's surface, and then textures and patterns of the fabric will be mapped into this deformation.
The cameras will capture the deformation under the help of reflective membrane and the illumination from LEDs. As a result, a sequence of tactile images with different locations are collected during rolling, which can be implemented to reconstruct the surface of the whole fabric.


Compared to the other optical tactile sensors discussed in Section~\ref{relatedwork}, we employ a cylindrical shape for the sensor instead of a flat surface which can roll over the object surface to collect tactile features. The camera is mounted in the center of the cylindrical tube, and ball bearings are used at the two sides of the sensor to make the camera point towards a fixed angle so that it can capture the tactile features while the sensor is rolling.

\subsection{Fabrication of the TouchRoller sensor}
\begin{figure}
	\centering
	\includegraphics[width=0.8\columnwidth]{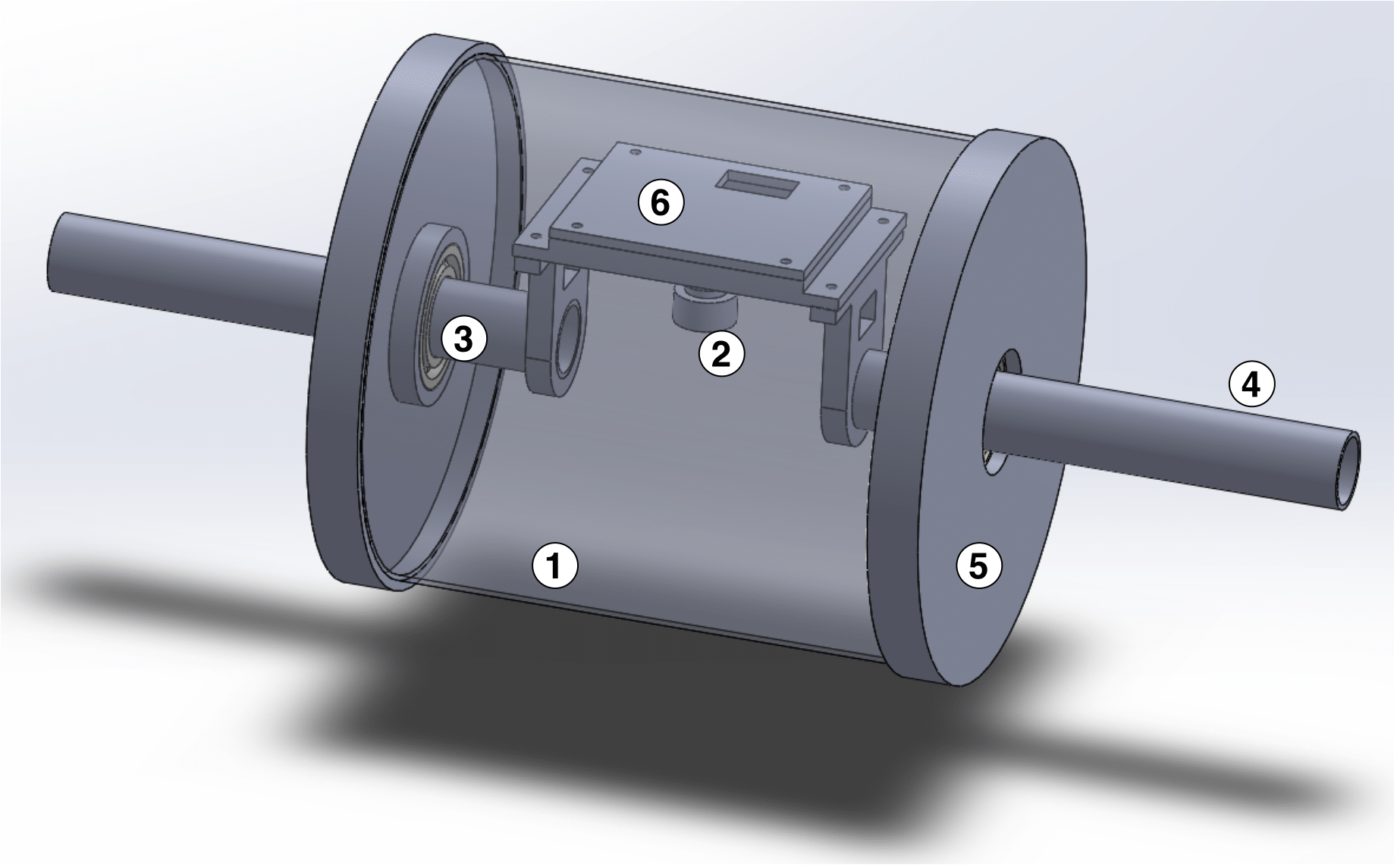}
	\caption{\textbf{\textit{The design of the TouchRoller tactile sensor.}} (1) The transparent cylindrical tube; (2) A webcam; (3) Ball bearings to make sure the camera point towards to the region in contact with the object; (4) Steel pipes; (5) Sleeves to fix the cylindrical tubes and pipes; (6) The supporting plate to place the camera.}
	\label{fig:shell}
\end{figure}
In this section, we will introduce the fabrication of our sensor, including the elastomer, reflective membrane, and the supporting parts of the sensor.

The elastomer, which serves as the contact-light media, is a crucial element in the optical tactile sensor. The elastomer is expected to be soft so that it is able to respond to the contact even with a small force, and it must also be optical clear so that the light can go through the elastomer with tiny refraction and attenuation. Moreover, in order to coat the elastomer on the cylindrical tube, the elastomer is required to be stretchable so that it can closely adhere to the surface of the cylindrical tube. 

In the fabrication of elastomer, we make use of the XP-565 from Silicones, Inc. and Slacker from Smooth-on Company as suggested in~\cite{gomes2020geltip}. XP-565 part A, XP-565 part B and Slacker are mixed with a ratio of $1:22:22$. In this mixture, the XP-565 part B can adjust the softness of elastomer, while the Slacker is used to increase the silicone tackiness.

The reflective membrane is used to reflect the light projected by LEDs with consistent reflectance. The main challenges of making the membrane are from the uneven painting that results in inconsistent reflectance, and cracks caused by the deformation of membrane, especially on a curved surface. To solve this problem, we mix the pigment in the dissolved silicone, and then paint this mixture on the surface of elastomer to form the membrane. Specifically, we use the Silver Cast Magic from Smooth-on Company and aluminium powder (mixed by 1:1) as our pigment. A small amount of XP-565 part A and part B as well as Slacker, the same proportion as for the elastomer, are mixed together with pigment at first, and then a silicone solvent is applied to dissolve this mixture to get the painting of the membrane. At last, we perform the painting using this mixture with air brush that can distribute the membrane evenly on the elastomer. As a result, the membrane will obtain a strong toughness and ductility to deform properly with different object surfaces.


The supporting parts, that are used to connect all elements and build up the sensor, consist of an off-the-shelf transparent glass tube, a set of ball bearings, two steel pipes, two sleeves to fix the glass tube, a supporting plate to place the camera, and the LEDs. As shown in Fig.~\ref{fig:shell}, the steel pipes are connected with the supporting plate, and the pipes go through the sleeves while the sleeves are used to fix the glass tube. Moreover, ball bearings are implemented between the pipes and the sleeve to enable the supporting plate pointing at a fixed angle while the sensor rolls over an object surface. Finally, we place the camera, with a resolution of $640\times480$ and a $2.8 mm$ focal length, on the supporting plate, a set of LEDs are placed beside the camera to illuminate the space inside the sensor. 
In this work, we use the 3D printer Anycubic i3 Mega to make the customised parts, including the sleeves and the supporting plate. Overall, our TouchRoller sensor has a weight of 420 grams, with a diameter of $100 mm$ and a length of $100 mm$. All the 3D printed models can be found at \url{https://github.com/3PTelephant/TouchRoller} 


\subsection{Surface Projection} \label{projection}
\begin{figure}
	\centering
	\includegraphics[width=0.8\columnwidth]{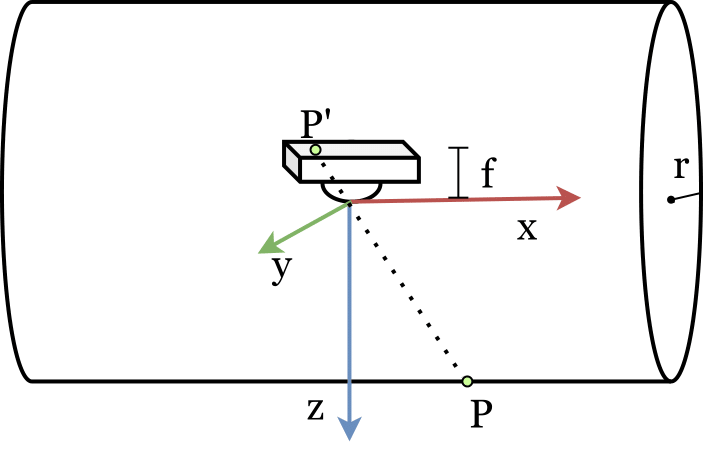}
	\caption{\textbf{\textit{The geometrical model of the TouchRoller tactile sensor.}} The sensor is modeled as a cylinder with a radius of $r$. A camera with focal length $f$ is placed in the center of the cylinder, and a point $P$ on the surface of the cylinder is mapped to $P'$ in the tactile image.}
	\label{fig:projection123}
\end{figure}

In the robotic inspection, it is crucial for the robot to know the location of the target object by the tactile feedback. Different from the GelSight sensor of a flat surface, our proposed rolling tactile sensor has a curved surface and it is necessary to determine the contact location from the tactile image collected by our sensor. To this end, we derive the location of contact on the surface of our sensor, given the contact pixels detected in the tactile image. 


As illustrated in Fig.~\ref{fig:projection123}, we model the surface of our TouchRoller sensor as a cylinder with radius $r$. The $x$-axis coincides with the central axis of the cylinder, and $z$-axis is vertically downward, with referential origin $(0,0,0)$ placed at the center of the cylinder. Therefore, the surface point $(x,y,z)$ of the sensor satisfies the following equation:
\begin{equation}\label{a}
y^{2}+z^{2}=r^{2}
\end{equation}

We model the camera as a pinhole camera model oriented to $z$-axis, with optical center of lens placed at referential origin. 
The transformation of a contact point $P=[X_{w}, Y_{w}, Z_{w}]^{T}$ to the pixel $P'=[u, v]^{T}$ in the tactile image can be represented~\cite{szeliski2010computer} as:
\begin{equation}\label{b}
\lambda\left[\begin{array}{l}
u \\
v \\
1
\end{array}\right]=K
\left[\begin{array}{cc}
R & t \\
0 & 1
\end{array}\right]\left[\begin{array}{c}
X_{w} \\
Y_{w} \\
Z_{w} \\
1
\end{array}\right]
\end{equation}
where both $P$, $P'$ are denoted in homogeneous coordinates, $\lambda$ is a scale factor for the image point, $K$ is the matrix of the intrinsic parameters of the camera, $R$ and $t$ represent the rotation and translation of the camera respectively. Firstly, $P$ is transformed by the camera's extrinsic matrix that gives the transformation between camera coordinate and world coordinate, and then mapped by the intrinsic matrix $K$ to get the pixel in the tactile image. 
In the camera's intrinsic matrix $K$:
\begin{equation}
K=\left[\begin{array}{llll}
\frac {f}{d_{x}} & 0 & u_{0} & 0 \\
0 & \frac {f}{d_{y}} & v_{0} & 0 \\
0 & 0 & 1 & 0
\end{array}\right]
\end{equation}
$f$ represents the focal length of the camera; $d_{x}$ and $d_{y}$ denote the pixel size; $(u_{0}, v_{0})$ is the center point of the tactile image.

As the tactile sensor rolls over a surface, we wish the camera oriented to the $z$-axis with its optical center of lens at the center of the cylinder. However, the camera may rotate while rolling, and the location can change due to the rotation. Considering the error in other orientation is very small that it can be ignored in our sensor, we only focus on the rotation on the $x$-axis and the translation on the $z$ axis in the extrinsic matrix. 
Specifically, $R$ and $t$ can be represented as follows:
\begin{equation}\label{c}
R=\left[\begin{array}{ccc}
1 & 0 & 0 \\
0 & \cos \theta & \sin \theta \\
0 & -\sin \theta & \cos \theta
\end{array}\right]
\end{equation}
\begin{equation}\label{d}
t=[ 0, 0, d ]^{T}
\end{equation}
where $\theta$ represents the rotation angle, and $d$ denotes the length of the vertical translation. By combining all the above equations, we can get:

\begin{equation}\label{bigeq}
\left\{\begin{array}{l}
\lambda u=X_{w}f_{x}-Y_{w}u_{0}\sin\theta+Z_{w}u_{0}\cos\theta+u_{0}d  \\
\lambda v=Y_{w}(f_{y}\cos\theta-v_{0}\sin\theta)+\\
\qquad\ {Z_{w}(f_{y}\sin\theta+v_{0}\cos\theta)+v_{0}d}\\
\lambda =-Y_{w}\sin\theta+Z_{w}\cos\theta+d\\
r^{2} = Y_{w}^{2}+Z_{w}^{2}
\end{array}\right.
\end{equation}
where the normalized focal length $f_{x}$ and $f_{y}$ denotes the $\frac {f}{d_{x}}$ and $\frac {f}{d_{y}}$ respectively. As a result, we can calculate the surface point $P=[X_{w}, Y_{w}, Z_{w}]^{T}$ according to the known $P'=[u, v]^{T}$ in tactile image using the Eq.~\ref{bigeq}.

\subsection{Calibration}\label{cali}
The sensor calibration aims to find both intrinsic parameter of this camera and the exact transformation matrix between the camera coordinate and the world coordinates. From Sec~\ref{projection}, we can map the pixel positions in the TouchRoller images to the world coordinates. However, the camera could not be absolutely vertical down and centered in the tube due to the error of installation and construction. Hence, the calibration is mandatory for precise measurement.

\begin{figure}
	\centering
\includegraphics[width=0.8\columnwidth]{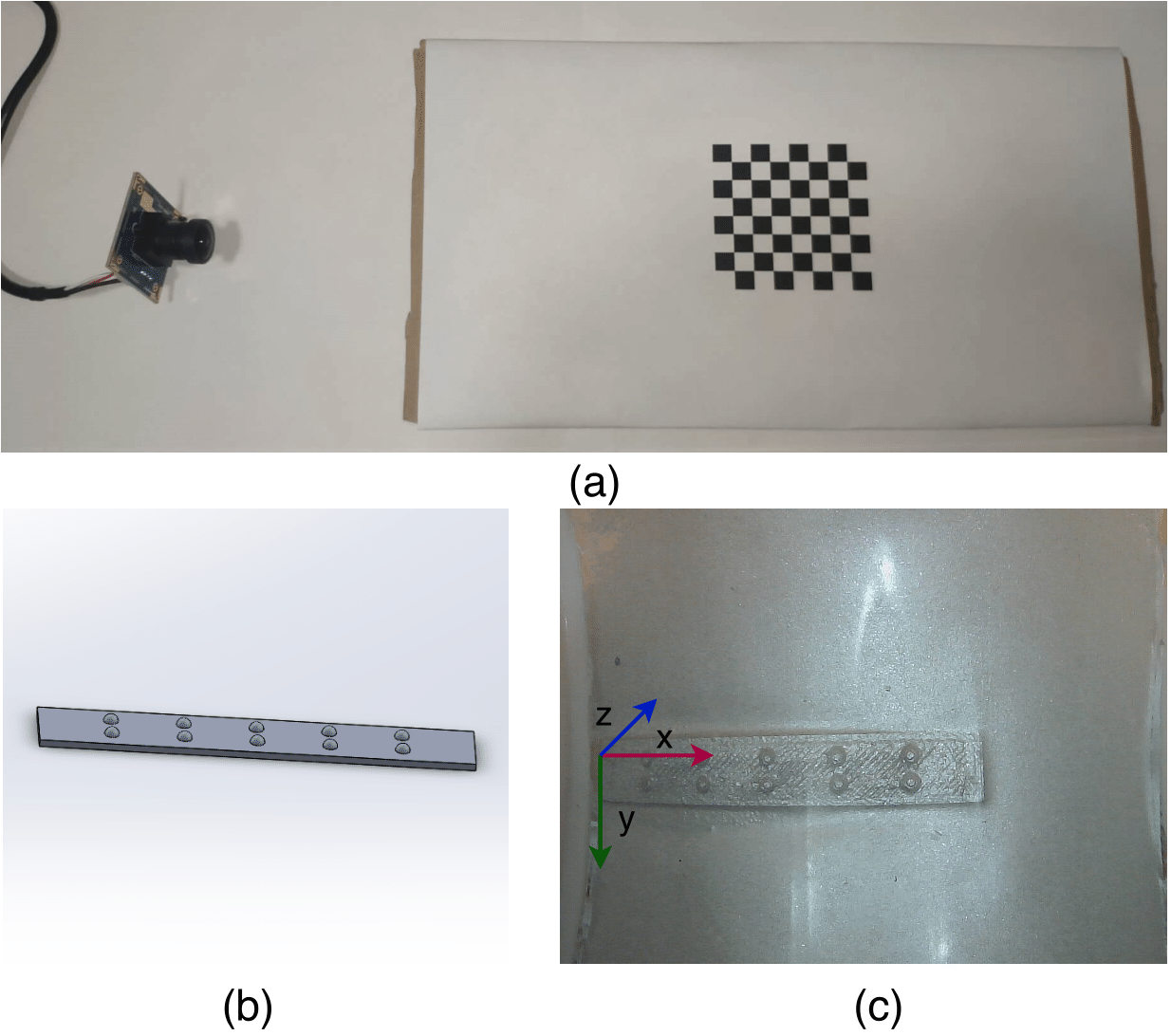}
	\caption{\textbf{\textit{The calibration of the TouchRoller sensor.}} (a) A $7\times7$ chessboard is used to calculate the intrinsic matrix of the camera. (b) A 3D printed solid with an array of $2\times5$ hemispheres is employed to estimate $R$ and $t$. (c) The corresponding tactile image when the sensor presses on the 3D printed solid shown in (b).   }
	\label{fig:calibration}
\end{figure}

Firstly, the intrinsic parameter of this camera is estimated with the widely used camera calibration method~\cite{zhang2000flexible}, as shown in Fig.~\ref{fig:calibration} (a). Then we use a small 3D-printed rectangle frame, as illustrated in Fig.~\ref{fig:calibration} (b), which consists of $2\times5$ hemispheres with a radius of 1 mm to estimate $\theta$ and $d$ in Eq.~\ref{bigeq}. During the calibration, we place our sensor on the surface of the rectangular frame and record a set of tactile images, with one example shown in Fig.~\ref{fig:calibration} (c). 

By using multiple image processing methods including 
binarization and morphological filtering, we can obtain the center of those hemispheres in the image space. Moreover, we define a reference coordinate in the rectangle frame as is shown in Fig.~\ref{fig:calibration} (c). With the known size of the frame and detected hemispheres centers, we first use the Iterative PnP solver~\cite{malik2002robust} to calculate the transformation matrix between the camera coordinates and the reference coordinates by minimizing the reprojection error in the pixel plane, and then get the rotation angle $\theta_i$ and translation error $t_i$ for each tactile image. Finally, we determine $\theta$ and $t$ with the average of all estimated $\theta_i$ and $t_i$. 

\section{Experiments and Analysis}
In this section, we report the results obtained on the carried out set of experiments to validate the design of our sensor. Firstly, a \textit{Surface Mapping} experiment is carried, which shows that the sensor can effectively be used to assess entire surfaces with a single linear rolling motion. Then, the \textit{Contact Localization}  experiment shows that contact points can be effectively localised through the proposed surface projection method. Finally, the perception efficiency is demonstrated through \textit{Comparison with other optical tactile sensors}.

\subsection{Surface Mapping }

We experiment mapping the texture of a $\SI{8}{\cm} \times \SI{11}{\cm}$ piece of fabric using three different speeds: Slow~(\SI{15}{\second}), Medium~(\SI{10}{\second}) and Fast~(\SI{5}{\second}). As shown in Fig.~\ref{fig:scene}, a video is captured by the internal camera of the sensor as it is rolled over the textured fabric for one full revolution. Thin centered patches corresponding to the region in contact, with a height of $70$ pixels, are then cropped from the captured video frames and stitched into one single tactile map. Each patch can be considered approximately flat, orthogonal to the camera sensor and, due to the design of the sensor, all captured at the same distance, $r$.  



\begin{figure}[h]
\centering
\includegraphics[width=0.9\columnwidth]{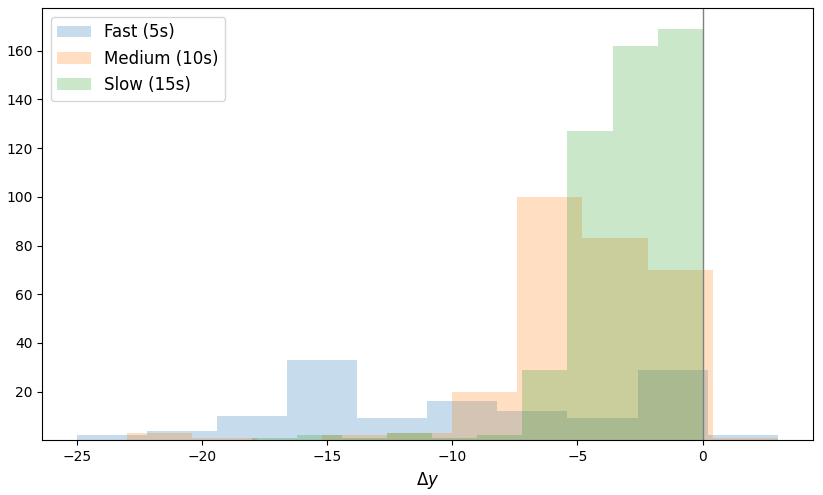}
\caption{Histograms of chosen $\Delta y$ for the three motion speeds. }
\label{fig:loss_error}
\end{figure}

As such, stitching the entire map requires only the finding of the $\Delta y$ translations, for every pair of consecutive frames, that minimizes the Mean Absolute Error (MAE) between two patches in the pair. $\Delta y$ is searched in $[-25, 25]$.  As expected, shown in Fig.~\ref{fig:loss_error} (the negative sign is representative of the direction of the motion, i.e., bottom-up), $\lvert \Delta y \lvert$ is directly proportional to the speed of the sensor.


For a quantitative analysis, the obtained map is compared with a gray-scale top-down view of the same textured fabric. To minimize the differences due to the two cameras viewing angles, distances to the surface and horizontal alignment, two pairs of points in the real and virtual frames are selected and constrained to fall in two corresponding vertical lines. A third pair of points is also derived such that the 3 points in each image form a right isosceles triangle. The OpenCV \textit{getAffineTransform} and \textit{warpAffine} functions are then applied to find the Affine Transformation between the tactile map and the gray-scale top-down view. This constrained alignment ensures that the spatial proportionality of the generated tactile map is preserved.

\begin{figure}[h]
\centering
\includegraphics[width=1\columnwidth]{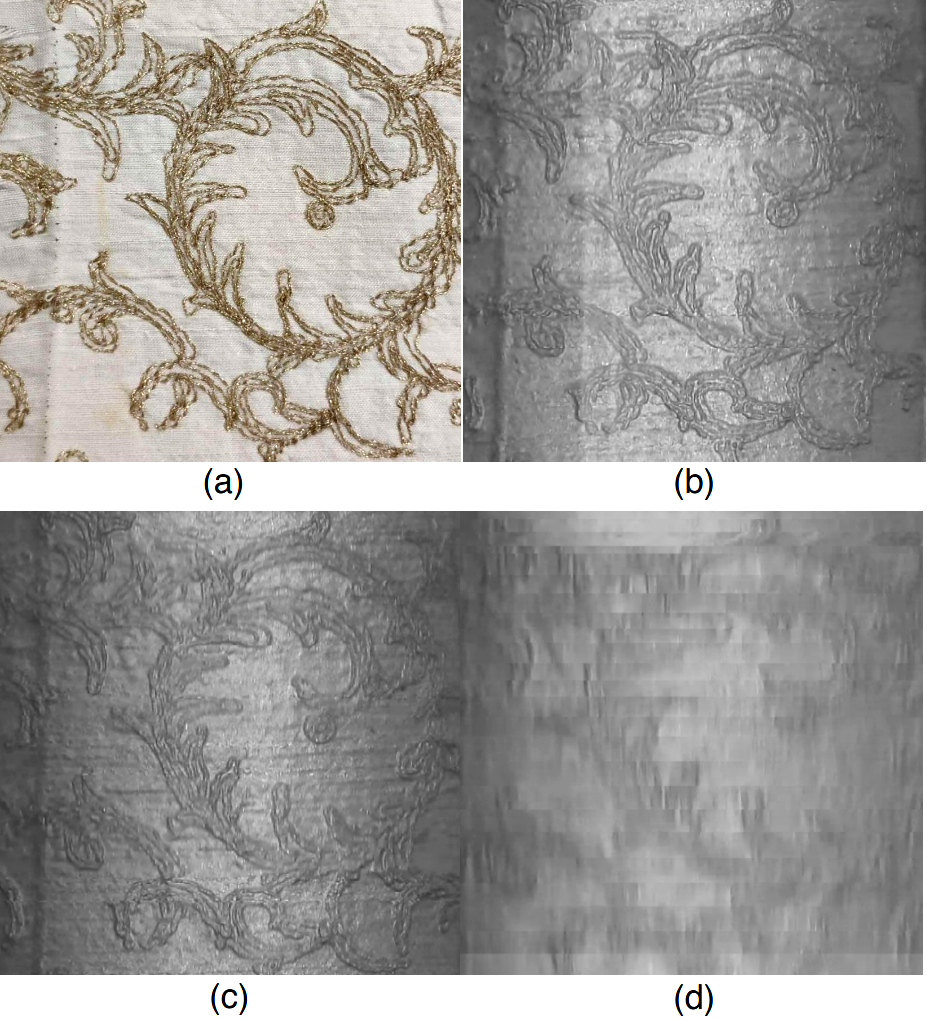}
\caption{\textbf{\textit{Mapped surface.}} (a) the visual top-down view, (b) (c) and (d) representative samples of the tactile maps obtained with the corresponding Slow~(\SI{15}{\second}), Medium~(\SI{10}{\second}) and Fast~(\SI{5}{\second}) motions.}
\label{fig:mapped_surface}
\end{figure}

Three evaluation metrics are chosen to assess the quality of the obtained tactile map: Structural Similarity (SSIM)~\cite{ssim}, Peak Signal-to-Noise Ratio (PSNR) and Mean Absolute Error (MAE), however, as shown in Table~\ref{table:alignment_errors}, the metrics fail to capture the poorest quality of the Fast mapping, and the failure may come from the difference between visual and tactile modalities. As shown in Fig.~\ref{fig:mapped_surface}, while the Fast motion clearly shows signs of motion blur, the two Medium and Slow motions appear well focused. As such, from this experiment it can be concluded that as long modest velocities are used (around \SI{1}{\cm/\second}) and therefore the proposed \textit{TouchRoller} sensor can be used to assess large surfaces rapidly.

\begin{table}
\centering
\caption{Comparison of the captured tactile map with the corresponding textured fabric top-view, for the three motion velocities: Slow~(\SI{15}{\second}), Medium~(\SI{10}{\second}) and Fast~(\SI{5}{\second}).}
\def\arraystretch{1.5}
\scalebox{0.85}{
\begin{tabular}{  l | c | c | c } 
\hline
 & SSIM & PSNR & MAE \\
\hhline{=|===}
\small{Slow   (\SI{15}{\second})}  
& $0.291\pm0.010$ & $15.09\pm0.13$ & $15.03\pm0.28$\%  \\
\hline
\small{Medium (\SI{10}{\second})  }
& $0.305\pm0.010$ & $14.41\pm0.71$ & $16.43\pm1.51$\%  \\
\hline
\small{Fast  (\SI{5}{\second})} 
& $\boldsymbol{0.330\pm0.007}$ & $\boldsymbol{15.59\pm0.94}$ & $\boldsymbol{14.04\pm2.04\%}$  \\
\hline
\end{tabular}}
\label{table:alignment_errors}
\end{table}

\subsection{Contact Localization }
Apart from data collection, it is also important for the robot to understand the contact location so that the robot is able to adjust pose to explore the object actively.
To this end, we use a 3D printed solid to tap on the surface of our sensor with different positions and derive the location in the world coordinates using the tactile image from our sensor. As a result, the localisation error can be measured between the derived location and the real contact location of the object.

\begin{figure}[h]
\begin{center}
   \includegraphics[width=0.9\columnwidth]{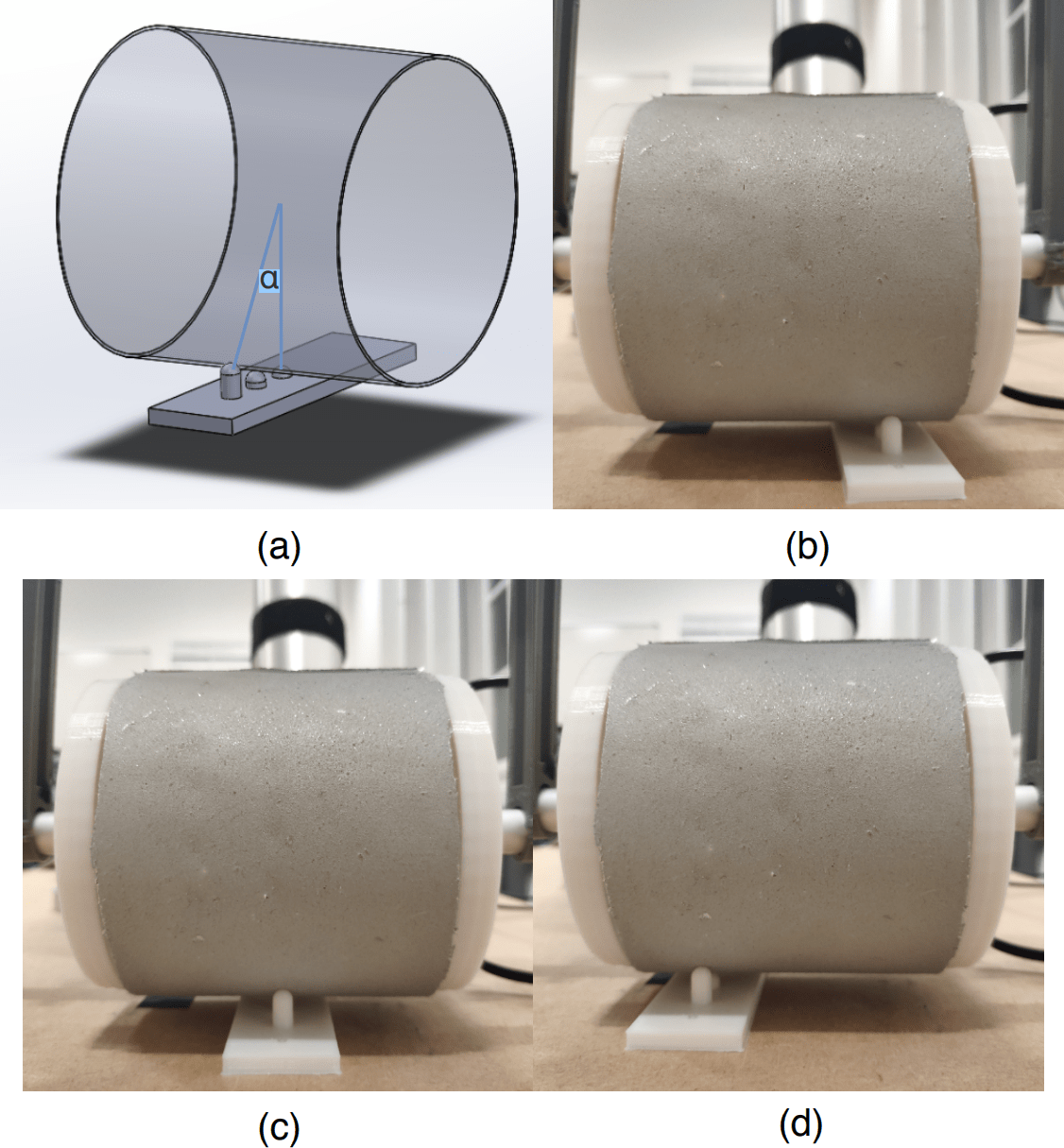}    
\end{center}
   \caption{\textbf{\textit{Contact localization}} (a) A 3D printed solid with three sticks of different heights is tapped vertically to the sensor ($\alpha$ is the central angles according to the heights). (b) (c) (d): The solid is also tapped on different locations of the sensor.  }
\label{fig:contactlocalization}
\end{figure}

As illustrated in Fig.~\ref{fig:contactlocalization} (a), a small solid with three sticks is printed with different heights. When we place the solid vertical to $x$-axis from both the front and back sides, the specific heights correspond to different central angles between the vertical direction and the connection of contacted points with the center of the circle (a certain section of the cylinder), i.e.,  $-\pi /6, -\pi/12, 0, \pi /12, \pi/6$ respectively. As shown in Fig.~\ref{fig:contactlocalization} (b) (c) (d), the solid is also tapped on the sensor along the $x$-axis on different locations, i.e, $1/4, 1/2, 3/4$ length of the sensor.

To determine the contact location in the tactile image, we design an extraction method. Firstly, the tactile image including the contact regions is grey-scaled and Gaussian blurred, then binarized using a proper threshold. Next, we implement the $FindContours$ function from OpenCV on the binarized image to determine the contact regions. At last, we make use of the $moments$ function from OpenCV to calculate the central points of each contact region, and these central points represent the contact locations in experiments.

As we have calculated the contact point in the tactile image, it is able for us to estimate the contact location using the surface projection method (detailed in Section~\ref{projection}). The sensor is first calibrated after being mounted onto the robot arm, and then the intrinsic matrix $K$ and extrinsic matrix which contains the $R$ and $t$ can be obtained. Since the radius of the cylinder $r$ is known, the contact location $P = [X_{w}, Y_{w}, Z_{w}]^{T}$ can be estimated using Eq.~\ref{bigeq}, with the detected contact point $P' = [u, v]^{T}$ in the tactile image.
\begin{table}
	\centering
\def\arraystretch{1.5}

		\caption{Euclidean distances between the estimated and the real contact locations (expressed in millimeters).}
		\label{tab:eudist}
		\begin{tabular}{c | c | c |c  }
			\hline
			Location &1/4 length &1/2 length& 3/4 length \\
			\hhline{=|=|=|=}
			$-\pi/6$ &9.64$\pm 0.09$&11.13$\pm 0.06$&13.75$\pm 1.89$ \\
			\hline 
            $-\pi/12$ &7.53$\pm 0.19$&4.50$\pm 0.08$&8.26$\pm 1.46$ \\
			\hline
			$0$ &$\boldsymbol{5.00\pm 0.59}$&$\boldsymbol{2.63\pm 0.74}$&6.89$\pm 0.79$\\
			\hline
			$\pi/12$&6.42$\pm 0.32$&4.06$\pm 0.26$&$\boldsymbol{6.66\pm 0.53}$\\
			\hline
			$\pi/6$ &8.83$\pm 0.10$&8.58$\pm 0.19$&12.34$\pm 0.21$\\
			\hline
		\end{tabular}
\end{table}

As shown in Table~\ref{tab:eudist}, the Euclidean distances between the estimated and the real contact location are computed, including both the mean and the standard deviation of the distance. We can find that the center of the sensing area has the smallest error, around $2.63mm$, compared with other regions. The error becomes larger while sensing area is further from the center. The error may arise from multiple factors: the calibration of the camera exists errors; the detection of the contact point in the tactile image is not precise enough; the thickness of the elastomer is uneven, and therefore the sensor is not a perfect cylinder.


\begin{figure}[h]
\begin{center}
   \includegraphics[width=0.9\columnwidth]{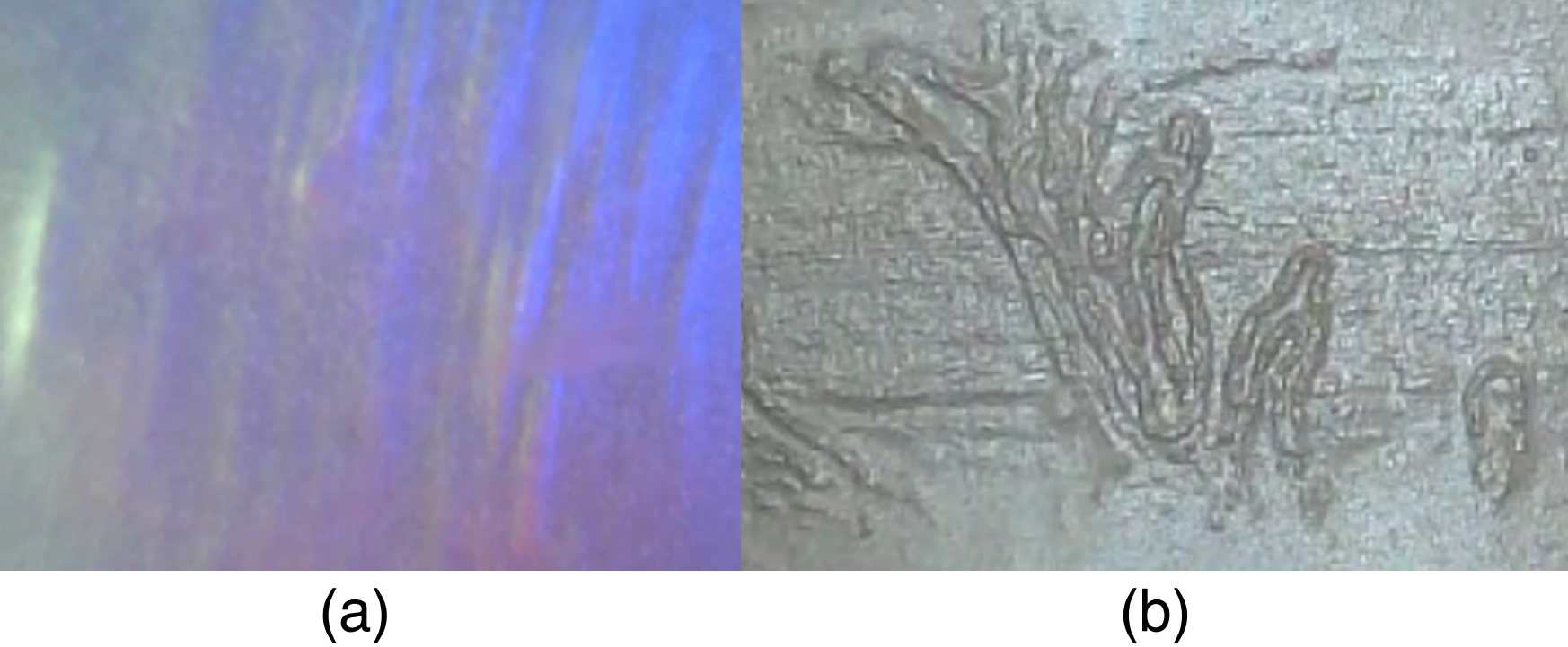}
\end{center}
   \caption{(a) The blurred tactile image collected by the GelSight while slipping on the fabric. (b) The clear tactile image by TouchRoller while rolling over the same piece of fabric.   }
\label{fig:compare}
\end{figure}

\subsection{Comparison with other optical tactile sensors}
In this section, our proposed sensor is compared with an optical tactile sensor of a flat surface, i.e., a GelSight sensor, in terms of the time consumption and definition of tactile features during the data collection.
The GelSight sensor is always used to press against the object with flat sensing area. 
If we slide the GelSight over an object, the tactile image will be blurred due to the unstable contact (shown in Fig.~\ref{fig:compare} (a)).
Hence, data collection by the GelSight requires a series of actions: pressing, lifting up and shifting to another region, and repeat again, which is very time consuming. 
In contrast, our proposed sensor is able to roll over the object with stable contacts, and it can generate clear tactile image (shown in Fig.~\ref{fig:compare} (b)) with rapid assessment of large surfaces. Specifically, to collect the tactile textures of a piece of fabric of $8cm\times11cm$, the GelSight sensor, with a sensing area of $1.6cm \times1.2cm$, requires at least $49$ pressing actions to cover the fabric, while the TouchRoller only need to roll over the fabric once. If we use both sensors to collect clear tactile images that cover the same piece of the fabric with a Universal Robots UR5 robot arm, the GelSight sensor takes about $196s$ while the TouchRoller only uses $10s$, which promotes the sampling efficiency significantly. 


\section{Conclusions}
In this paper, we introduce a novel design for a rolling optical tactile sensor \textit{TouchRoller} that has a cylindrical shape and can roll over an object surface to improve the efficiency of data collection. Our design demonstrates faster assessment of large surfaces by rolling the sensor over the surface of objects compared with other optical tactile sensors. Our experiments showed that the sensor can be used to reconstruct the surface of an object effectively and the contact on the sensor surface can be localised with a high accuracy. Our TouchRoller tactile sensor can be applied in fast surface inspection including crack detection in structural health monitoring. In the future work, we will minimize the size of the sensor so that it can be mounted onto the robot grippers and used in applications such as robot grasping.

{\small
	\bibliographystyle{ieeetr}
	\bibliography{reference.bib}
}

\end{document}